# Confidence Inference in Bayesian Networks

Jian Cheng & Marek J. Druzdzel*
Decision Systems Laboratory
School of Information Sciences and Intelligent Systems Program
University of Pittsburgh, Pittsburgh, PA 15260
{jcheng,marek}@sis.pitt.edu

## Abstract

We present two sampling algorithms for probabilistic confidence inference in Bayesian networks. These two algorithms (we call them AIS-BN-$\mu$ and AIS-BN-$\sigma$ algorithms) guarantee that estimates of posterior probabilities are with a given probability within a desired precision bound. Our algorithms are based on recent advances in sampling algorithms for (1) estimating the mean of bounded random variables and (2) adaptive importance sampling in Bayesian networks. In addition to a simple stopping rule for sampling that they provide, the AIS-BN-$\mu$ and AIS-BN-$\sigma$ algorithms are capable of guiding the learning process in the AIS-BN algorithm. An empirical evaluation of the proposed algorithms shows excellent performance, even for very unlikely evidence.

## 1 Introduction

The main application of stochastic sampling algorithms in Bayesian networks is inference in very large networks in which exact methods are intractable. Stochastic sampling algorithms essentially trade off precision for computation — sample generation can be interrupted at any time yielding an approximate answer. While absolute precision is seldom critical, it is often useful to know roughly how close the answer is to the exact answer or, in other words, what is the confidence interval around the computed result.

The best existing algorithms that address this problem are the *bounded variance algorithm* [Dagum and Luby, 1997] and the AA algorithm [Dagum *et al.*, 2000]. The test results reported in Pradhan and Dagum [1996] show that these algorithms work well when the probability of evidence is not too small. However, as simple tests that we conducted showed, in very large networks, especially when several observations have been made and the probability of evidence is very small, these algorithms usually require a prohibitive number of samples to satisfy the requirement. One of the reasons for this is that these algorithms are based on the likelihood weighting algorithm [Fung and Chang, 1989, Shachter and Peot, 1989], which suffers from the problem of mismatch between the optimal and the actually used importance function (see [Cheng and Druzdzel, 2000] for an in-depth discussion of this problem). Another problem is that the confidence intervals calculated by these algorithms are not tight enough. Recent advances in simulation algorithms, notably the AIS-BN algorithm [Cheng and Druzdzel, 2000] and stopping rules [Cheng, 2001], address both of these problems.

*Currently with ReasonEdge Technologies, Pte, Ltd, 438 Alexandra Road, #03-01A Alexandra Point, Singapore 119-958, Republic of Singapore, mjdruzdzel@reasonedge.com.

In this paper, we combine the AIS-BN simulation algorithm with the new stopping rules to yield two sampling algorithms that perform well in very large networks. Essentially, our approach is to use the new stopping rules in the AIS-BN algorithm to guide the process of learning the importance function. After each learning step, we use the SA-$\mu$ or SA-$\sigma$ algorithm [Cheng, 2001] to produce the estimated number of samples that is needed to achieve a required precision. SA-$\mu$ and SA-$\sigma$ algorithms are currently the best known distribution-independent algorithms to estimate the mean and they require a relatively small number of samples. In addition to estimating the number of samples needed to achieve a desired precision, the resulting AIS-BN-$\mu$ and AIS-BN-$\sigma$ algorithms react to situations when the number of samples is prohibitive. In our approach, the algorithms use heuristic methods to modify the importance function combined with a restart, whereby typically the second try solves the problem of prohibitive computation.

In the following discussion, capital letters, such as $A$, $B$, or $C$ will denote multiple-valued, discrete random variables. Bold capital letters, such as $\mathbf{A}$, $\mathbf{B}$, or $\mathbf{C}$, denote sets of variables. $\mathbf{E}$ will denotes the set of evidence variables. Lower case letters $a$, $b$, $c$ denote particular instantiations of variables $A$, $B$, and $C$ respectively. Bold lower case letters, such as $\mathbf{a}$, $\mathbf{b}$, $\mathbf{c}$, and $\mathbf{e}$, denote particular instantiations of $\mathbf{A}$, $\mathbf{B}$, $\mathbf{C}$, and $\mathbf{E}$ respectively. Pa($A$) denotes the parents of node $A$. Pr($\mathbf{X}$) denotes the network joint probability distribution. \ denotes set difference. Vertical bar, such as in Pa($\mathbf{A}$)$|_{\mathbf{E}=\mathbf{e}}$, denotes substitution of $\mathbf{e}$ for $\mathbf{E}$ in $\mathbf{A}$. $w^{(k)}$ or $\Pr^{(k)}$ denote a number or a function in stage $k$.

## 2 AIS-BN: Adaptive Importance Sampling for Bayesian Networks

Because familiarity with the AIS-BN algorithm will be helpful in understanding the current paper, we will briefly review its design. Readers interested in details are directed to the exposition in [Cheng and Druzdzel, 2000].

The AIS-BN algorithm is based on importance sampling in finite dimensional integrals. Using the structural advantages of Bayesian networks, it tries to reduce sampling variance by learning a sampling distribution $\Pr^{(i)}(\mathbf{X}\backslash\mathbf{E})$ that is as close as possible to the optimal importance sampling function. Since the sampling distributions are different in every updating step, the AIS-BN algorithm introduces different weights for samples generated at different learning stages. Our experimental results show that the AIS-BN algorithm can improve the convergence rate dramatically compared to other existing sampling algorithms. We observed typically two orders of magnitude improvement in precision of the results expressed by mean square error.

Suppose that the importance sampling function used in the AIS-BN algorithm is $\Pr'(\mathbf{X}\backslash\mathbf{E})$. By defining a random variable

$$Z(\mathbf{X}\backslash\mathbf{E}) := \frac{\Pr(\mathbf{X}\backslash\mathbf{E}, \mathbf{E}=\mathbf{e})}{\Pr'(\mathbf{X}\backslash\mathbf{E})}, \quad (1)$$

we obtain $Z(\mathbf{s})$, an unbiased estimate of $\Pr(\mathbf{E}=\mathbf{e})$. Here $\mathbf{s}$ is a random sample from $\Pr'(\mathbf{X}\backslash\mathbf{E})$.

The most important component of the AIS-BN algorithm is learning the importance function. The closer an importance function is to the optimal importance function, the smaller the required number of samples to satisfy the desired precision. The updating formula used by the AIS-BN algorithm is

$$\Pr^{(k+1)}(x_i|\text{pa}(X_i), \mathbf{e}) = \Pr^{(k)}(x_i|\text{pa}(X_i), \mathbf{e}) + \eta(k) \cdot [\Pr'(x_i|\text{pa}(X_i), \mathbf{e}) - \Pr^{(k)}(x_i|\text{pa}(X_i), \mathbf{e})],$$

where $\Pr^{(k+1)}(x_i|\text{pa}(X_i), \mathbf{e})$ is the updated conditional probability, $\Pr^{(k)}(x_i|\text{pa}(X_i), \mathbf{e})$ is the current sampling conditional probability, and $\Pr'(x_i|\text{pa}(X_i), \mathbf{e})$ is the estimated conditional probability based on current samples. The latter can be obtained by counting score sums corresponding to $\{x_i, \text{pa}(X_i), \mathbf{e}\}$. $\eta(k)$ is the rate of learning that influences directly the convergence speed. A good rate will let $\Pr^{(k+1)}(x_i|\text{pa}(X_i), \mathbf{e})$ converge to the destination function $\Pr(x_i|\text{pa}(X_i), \mathbf{e})$ quickly. Too small or too large $\eta(k)$ may lead to slow convergence. The analysis presented later in this paper will shed some light on the optimal choice of the convergence rate.

The weighting function $w^{(k)}$ determines how estimates from the different sampling distributions are combined and is another parameter that needs to be chosen in the AIS-BN algorithm. Although in [Cheng and Druzdzel, 2000] we recommended choosing $w^{(k)} \propto 1/\widehat{\sigma}^{(k)}$, where $\widehat{\sigma}^{(k)}$ is the estimated standard deviation at Stage $k$, based on the new stopping rules in this paper we will propose an improved weighting scheme.

## 3 Preliminary Analysis

Before discussing stopping rules, we first review some important approximation concepts that will be used in this paper. By *absolute approximation* we mean an estimate $\hat{\mu}$ of $\mu$ that satisfies $|\hat{\mu} - \mu| \leq \varepsilon_a$. *Relative approximation* is an estimate $\hat{\mu}$ of $\mu$ that satisfies $\frac{|\hat{\mu}-\mu|}{\mu} \leq \varepsilon_r$. $(\varepsilon_a, \delta)$ *absolute approximation* is an estimate $\hat{\mu}$ of $\mu$ that satisfies $\Pr(|\hat{\mu} - \mu| \leq \varepsilon_a) \geq 1 - \delta$. $(\varepsilon_r, \delta)$ *relative approximation:* is an estimate $\hat{\mu}$ of $\mu$ that satisfies $\Pr(|\hat{\mu} - \mu| \leq \varepsilon_r \mu) \geq 1 - \delta$. We use $\varepsilon_a$ to denote absolute error, $\varepsilon_r$ to denote relative error, and $1 - \delta$ to denote confidence level. One can see that, for $\mu \neq 0$, $\varepsilon_a = \varepsilon_r \cdot \mu$. We are only interested in the case where $0 < \varepsilon_r, \delta < 1$. When the range of $\mu$ is unknown, we are more interested in the relative approximation than absolute approximation.

In computing a posterior probability $\Pr(\mathbf{a}|\mathbf{e})$ by simulation, the values of $\Pr(\mathbf{a}, \mathbf{e})$ and $\Pr(\mathbf{e})$ are estimated separately. Subsequently, the definition of the conditional probability, $\Pr(\mathbf{a}|\mathbf{e}) = \Pr(\mathbf{a}, \mathbf{e})/\Pr(\mathbf{e})$, yields the result. If we use absolute approximations for $\Pr(\mathbf{a}, \mathbf{e})$ and $\Pr(\mathbf{e})$, it is difficult to give an error estimate of $\Pr(\mathbf{a}|\mathbf{e})$. However, if we know that relative approximation and the confidence level for both $\Pr(\mathbf{a}, \mathbf{e})$ and $\Pr(\mathbf{e})$ are $\varepsilon_r$ and $1-\delta$ respectively, we can get a relative approximation for $\Pr(\mathbf{a}|\mathbf{e})$

$$\frac{-2\varepsilon_r}{1+\varepsilon_r} \leq \frac{\Pr'(\mathbf{a}|\mathbf{e}) - \Pr(\mathbf{a}|\mathbf{e})}{\Pr(\mathbf{a}|\mathbf{e})} \leq \frac{2\varepsilon_r}{1-\varepsilon_r} \quad (2)$$

with the confidence level of at least $1 - 2\delta$. Both estimates are conservative.

Stopping rules give the number of samples $N$ that guarantees to achieve the specified $(\varepsilon, \delta)$ approximation of $\mu_Z$. Several researchers have investigated stopping rules in the context of stochastic sampling algorithms, e.g., [Chavez and Cooper, 1990, Dagum and Horvitz, 1993, Dagum et al., 2000, Cheng, 2001]. As far as we know, currently the tightest estimates are those reported in [Cheng, 2001], based on the following two theorems.

**Theorem 3.1** Let $Z_1$, $Z_2$, ..., $Z_N$ be independent and identically distributed random variables with $E(Z_i) = \mu_Z$, $0 \leq Z_i \leq b$, $i = 1, \ldots, N$. If $0 < \varepsilon_r < \min(1, b/\mu_Z - 1)$ and

$$N \geq \frac{b}{\mu_Z} \cdot \frac{1}{(1+\varepsilon_r)\ln(1+\varepsilon_r) - \varepsilon_r} \ln \frac{2}{\delta}, \quad (3)$$

then $\overline{Z} = (Z_1 + \ldots + Z_N)/N$ is an $(\varepsilon_r, \delta)$ relative approximation of $\mu_Z$.

**Theorem 3.2** Let $Z_1$, $Z_2$, ..., $Z_N$ be independent and identically distributed random variables with $E(Z_i) = \mu_Z$, $Var(Z_i) = \sigma_Z^2$, $0 \leq Z_i \leq b$, $i = 1, \ldots, N$. If $0 < \varepsilon_r < 1$ and

$$N \geq \frac{b}{\mu_Z} \cdot \frac{1}{\varepsilon_r[(1+\frac{\sigma_Z^2}{b\varepsilon_r\mu_Z})\ln(1+\frac{b\varepsilon_r\mu_Z}{\sigma_Z^2}) - 1]} \ln \frac{2}{\delta}, \quad (4)$$

then $\overline{Z} = (Z_1 + \ldots + Z_N)/N$ is an $(\varepsilon_r, \delta)$ relative approximation of $\mu_Z$.

Theorems 3.1 and 3.2 form the basis of the theoretical analysis presented in this paper. Notice that the main difference between Theorem 3.1 and Theorem 3.2 is that the former does not require the knowledge of variance.

From Theorem 3.1 we can see that if there are two variables that have the same mean but a different bound $b$, the variable with a smaller bound requires a smaller minimum number of samples. For a fixed bound $b$, according to Theorem 3.2, it is not difficult to prove that the minimum required number of samples is a strictly increasing function of variance $\sigma_Z^2$. So, if there exists a way to define a variable that has the same mean as the known variable but has a smaller bound and a smaller variance, then this will lead to a decrease in the minimum required number of samples. Adaptive importance sampling is based on this idea. It focuses on finding a sampling distribution $Pr'(\mathbf{X}\backslash\mathbf{E})$ in equation (1) that can significantly decrease the bound and variance of $Z(\mathbf{X}\backslash\mathbf{E})$. The judgment whether one sampling distribution $Pr'(\mathbf{X}\backslash\mathbf{E})$ is better than another can be made by comparing the minimum required number of samples $N$ obtained by means of inequalities (3) and (4).

The calculation of $N$ requires the exact value of the mean. This, however, is the value that we want to estimate and, hence, we cannot use the stopping rules directly. But based on the stopping rules, the SA-$\mu$ and SA-$\sigma$ algorithms [Cheng, 2001] circumvent this problem. These two algorithms guarantee that the sampling result $\tilde{\mu}_Z$ is an $(\varepsilon_r, \delta)$ relative approximation of $\mu_Z$. The mean number of samples in the SA-$\mu$ algorithm is very close to the requirement in Theorem 3.1 [Cheng, 2001].

While the maximum variance of a random variable is $(b - \mu_Z) \cdot \mu_Z$, the real variance can be much smaller. So, the algorithm based on the stopping rule with the knowledge of variance is almost always better than one without the knowledge of variance. We recommend using the SA-$\sigma$ algorithm even if the exact value of $\sigma$ is not known — a conservative estimate of $\sigma$ will still save much computation.

Let the tightest bound of a random variable be $t_b$. In case of the likelihood weighting algorithm, it is not difficult to get an upper bound on $t_b$. We define $u_i$ to be the largest value in the conditional probability table $\Pr(x_i|pa(X_i))$, excluding the values that are not consistent with observed evidence $\mathbf{e}$. The likelihood weighting algorithm corresponds to the following choice of the importance function

$$\Pr'(\mathbf{X}\backslash\mathbf{E}) = \prod_{X_i \notin \mathbf{E}} \Pr(X_i|pa(X_i)) \bigg|_{\mathbf{E=e}}.$$

As a result, we can get an upper bound on $Z(\mathbf{X}\backslash\mathbf{E})$

$$Z(\mathbf{X}\backslash\mathbf{E}) \leq \prod_{X_i \in \mathbf{E}} u_i \leq 1. \quad (5)$$

We should point out that $\prod_{X_i \in \mathbf{E}} u_i$ is not necessarily the best bound and the tightest bound $t_b$ can be several orders of magnitude smaller. For other kinds of sampling distributions $Pr'(\mathbf{X}\backslash\mathbf{E})$, there is no easy way to get a tighter bound, or the estimated bound is too crude to be used. As a matter of fact, calculation of the tightest bound $t_b$ is isomorphic to the Maximum A-Posteriori assigment problem (MAP) [Pearl, 1988]. MAP corresponds to calculating the largest value of $\Pr(\mathbf{X}\backslash\mathbf{E}, \mathbf{E=e})$ and the tightest bound $t_b$ corresponds to calculating the largest value of $Z(\mathbf{X}\backslash\mathbf{E})$. Since computing the MAP in Bayesian networks is NP-hard [Shimony, 1994], the value of $b$ has to be estimated in practice (note that we focus on inference in very large networks). In our algorithms, we will use for $b$ the largest random value in the samples that are generated by the sampling distribution $Pr'(\mathbf{X}\backslash\mathbf{E})$. In the SA-$\sigma$ algorithm, we also need the value of $\sigma_Z^2$. In case of a simulation algorithm, this value is impossible to obtain in advance but can be estimated from available samples, for example by $\tilde{\sigma}_Z^2 = 1/(N-1) \cdot (\sum_{j=1}^{N} Z_j^2 - N \cdot \overline{Z}^2)$, or by the technique addressed in [Fishman, 1995] to avoid possible numerical errors caused by the limited precision. Our experimental results, presented in Section 6, show that these approximations to $t_b$ and $\sigma_Z^2$ are reasonable.

Given the estimated values $\tilde{b}$, $\tilde{\sigma}_Z^2$, and $\tilde{\mu}_Z$, inequality (4) allows us to obtain an estimated minimum required number of samples $\tilde{N}$ for a given relative approximation of $\mu_Z$. $\tilde{N}$ can be used to judge whether one sampling distribution is better than another. The learning rate $\eta(k)$ and the weighting function $w^{(k)}$ can be also based on this number. With respect to the weighting function, if there are two sampling distributions, $\Pr^{(k)}(\mathbf{X} \backslash \mathbf{E})$ and $\Pr^{(k+1)}(\mathbf{X} \backslash \mathbf{E})$, and their corresponding estimated minimum required number of samples are $\tilde{N}^{(k)}$ and $\tilde{N}^{(k+1)}$, then the weighting function should satisfy $w^{(k+1)}/w^{(k)} = \tilde{N}^{(k)}/\tilde{N}^{(k+1)}$, since $\tilde{N}^{(k)}$ samples from $\Pr^{(k)}(\mathbf{X} \backslash \mathbf{E})$ will yield almost the same relative approximation of $\mu_Z$ as $\tilde{N}^{(k+1)}$ samples from $\Pr^{(k+1)}(\mathbf{X} \backslash \mathbf{E})$. We can also use this relationship to convert $l$ samples from $\Pr^{(k)}(\mathbf{X} \backslash \mathbf{E})$ to $l \cdot \tilde{N}^{(k+1)}/\tilde{N}^{(k)}$ samples in $\Pr^{(k+1)}(\mathbf{X} \backslash \mathbf{E})$. After normalizing the weighting function $w^{(k)}$, $w^{(k)}$ should satisfy $\sum_{k=1}^{n} w^{(k)} = 1$. Solving these equations, we get

$$w^{(k)} = \frac{1/\tilde{N}^{(k)}}{\sum_{l=1}^{n}(1/\tilde{N}^{(l)})}.$$

So the contribution of the estimated probability from the stage $k$ can be calculated as $Z_{\text{TScore}}/(l \cdot \tilde{N}^{(k)})$. To normalize this value, we divide it by $\sum_{l=1}^{n}(1/\tilde{N}^{(l)})$ in the final step of the algorithm (see Figure 1). We will discuss the adjustment of the learning rate $\eta(k)$ in the context of the empirical tests of our algorithms.

## 4 The AIS-BN-$\mu$ and AIS-BN-$\sigma$ Algorithms

Based on the analysis presented in the previous section, we propose an algorithm that combines the AIS-BN algorithm with the SA-$\sigma$ algorithm. To simplify the notation, we will call this algorithm AIS-BN-$\sigma$ (Figure 1). An algorithm that combines the AIS-BN algorithm with the SA-$\mu$ algorithm (AIS-BN-$\mu$) can be obtained following an analogous process.

In the AIS-BN-$\sigma$ algorithm, we need a function $\delta = f_\sigma(\delta_s, \varepsilon_r)$. Its definition and the table listing the relationship between $\delta_s$, $\delta$, and $\varepsilon_r$ can be found in [Cheng, 2001]. $\delta_s$ is very close to $\delta$ in the range of interest — when $\varepsilon_r \leq 0.01$, $\delta \approx \delta_s$.

The methods of initializing the importance function $\Pr^{(0)}(\mathbf{X} \backslash \mathbf{W})$, generating a sample according to

```
Input: (εr, δ) with 0 < εr < 1, 0 < δ < 1, the up-
dating interval l, a threshold value t < l, evidence
E = e, query states Aj = aj, j = 1, ..., m.

Output: μ̃j, j = 1, ..., m.

Procedure AIS-BN-σ
φe ← Estimate_Prob(E = e)
for j ← 1 to m
    w ← e ∪ aj
    φj ← Estimate_Prob(W = w)
    μ̃j ← φj/φe
end for

Function Estimate_Prob(W = w)
(Estimate the probability of a set of variables W
being equal to w: Pr(W = w).)
δs ← fσ⁻¹(δ, εr)
α ← 1/(εr·(1−εr)) · ln(2/δs)
γ ← 0, k ← 0, i ← 0, b̃ ← 0, ζ ← 0
ZTScore ← 0, ωTScore ← 0, ωsum ← 0
Initialize the importance function Pr⁽⁰⁾(X\W) us-
ing some heuristic methods
repeat
    si ← Generate a sample according to
           Pr⁽ᵏ⁾(X\W)
    ZiScore ← Pr(si, W = w)/Pr⁽ᵏ⁾(si)
    ZTScore ← ZTScore + ZiScore
    ζ ← ζ + Z²iScore
    if (b̃ < ZiScore) then b̃ ← ZiScore end if
    i ← i + 1
    if (i > t) then
        μ̃Z ← ZTScore/i
        σ̃²Z ← (ζ − i · μ̃²Z)/(i − 1)
        Ñ ← α · b̃/[(μ̃Z + σ̃²Z/(b̃εr)) ln(1 + b̃εrμ̃Z/σ̃²Z) − μ̃Z]
    end if
    if (i == l) then
        k ← k + 1
        Update the importance function
           Pr⁽ᵏ⁾(X\W)
        γ ← γ + l/Ñ
        ωTScore ← ωTScore + ZTScore/(l · Ñ)
        ωsum ← ωsum + 1/Ñ
        i ← 0, b̃ ← 0, ζ ← 0, ZTScore ← 0
    end if
until (i ≥ max(t, (1 − γ) · Ñ))
ωTScore ← ωTScore + ZTScore/(i · Ñ)
ωsum ← ωsum + 1/Ñ
return ωTScore/ωsum
```

Figure 1: The AIS-BN-$\sigma$ algorithm combining the AIS-BN algorithm with the SA-$\sigma$ algorithm.

$\Pr^{(k)}(\mathbf{X}\backslash\mathbf{W})$, and updating the importance function $\Pr^{(k)}(\mathbf{X}\backslash\mathbf{W})$ are discussed in [Cheng and Druzdzel, 2000]. The parameter $\gamma$ stands for the percentage of the samples that have been generated to satisfy the precision requirement. To avoid the situation in which the estimates of $\widetilde{b}$ and $\widetilde{\sigma}_Z^2$ are too far away from the exact value, we use a threshold value $t$ to make sure that the number of samples used to estimate $\widetilde{b}$ and $\widetilde{\sigma}_Z^2$ is sufficiently large. When $i$ is smaller than $t$ in a new stage, we can either skip the judgment $i \geq (1-\gamma) \cdot \widetilde{N}$ or use the previously estimated $\widetilde{N}^{(k-1)}$ to judge if the number of samples has satisfied our requirement (theoretically, $N^{(k-1)}$ should be larger than $N^{(k)}$). To facilitate the learning process for $\Pr(\mathbf{X}\backslash\{\mathbf{E}\cup\mathbf{A_j}\})$, $j = 1, \ldots, m$, we can adapt the final learned importance function $\Pr^{(k)}(\mathbf{X}\backslash\mathbf{E})$, which is obtained when we estimate the probability of evidence, to initialize the importance function $\Pr^{(0)}(\mathbf{X}\backslash\{\mathbf{E}\cup\mathbf{A_j}\})$. This method should lead to considerable savings.

There is a tradeoff between the time spent on sampling and the time spent on updating the importance function $\Pr^{(k)}(\mathbf{X}\backslash\mathbf{W})$. Several methods can be used to address this trade-off. One method is to focus on learning until its convergence becomes slow and then to sample from the learned importance function. Theoretically, the learning convergence can be judged by the minimum required number of samples, for which $\tilde{N}^{(k)}$ is a good proxy. Using only the samples that are generated after finishing the learning stage to estimate $\widetilde{b}$, $\widetilde{\sigma}_Z^2$, and $\tilde{N}^{(k)}$ avoids a possible error introduced by $\gamma$. The advantage of this method is that it facilitates obtaining good estimates of $\widetilde{b}$ and $\widetilde{\sigma}_Z^2$ and, at the same time, generate more samples. Another method is to interleave learning and sampling, but to let $t$ and $l$ be sufficiently large. The advantage of the latter method is that our importance distribution will converge to the target importance function that we want to learn. The disadvantage is that the estimates of $\widetilde{b}$ and $\widetilde{\sigma}_Z^2$ may not be sufficiently accurate and may introduce error into $\gamma$. The former method will generally generate more samples within the same amount of time. Since the importance sampling functions during the initial stages of learning will generally introduce large variance into the results (the estimates of $\widetilde{b}$, $\widetilde{\sigma}_Z^2$ and $\tilde{\mu}_Z$), we also suggest to focus purely on learning and to discard the samples in the first few stages of the algorithm.

There are various methods for initializing the function $\Pr^{(0)}(\mathbf{X}\backslash\mathbf{W})$ and there seems to be no general rule for choosing one method over another. But since based on available samples, we can get an estimated minimum required number of samples $\tilde{N}^{(k)}$, we can use this number to judge the initialization, along with the convergence. If after several updating stages, we still require a prohibitive number of samples, we can change the initialization method and try again.

We use the estimated $b$ and $\sigma_Z^2$ to calculate $N$, which inevitably introduces error. However, our experimental results show that the approximation is reasonable, because the algorithms are based on the worst-case scenarios in how they treat the inequalities. To guarantee the precision requirement, we can adopt an upper bound of $t_b$ into the AIS-BN-$\mu$ algorithm, such as to the likelihood weighting algorithm, we can use the bound in inequality (5) to guarantee the results. But usually, the difference between the actual value of $t_b$ and its upper bound is so large that we cannot afford the required number of samples using this method. Approximating $t_b$ is possibly the only method viable in practice.

Sometimes, we are also interested in the relative error or confidence level in a given stage of simulation. This can be calculated using inequalities (3) and (4).

## 5 Related Work

Dagum et al. [2000] proposed a stopping rule called Generalized Zero-One Estimator Theory. To let $\overline{Z}$ be an $(\varepsilon_r, \delta)$ relative approximation of $\mu_Z$ (assume $Z_i$ is in the interval $[0, 1]$), the required number of samples in the Generalized Zero-One Estimator Theory is

$$N \geq \frac{4\lambda\rho_Z}{\mu_Z^2 \varepsilon_r^2} \ln \frac{2}{\delta}, \qquad (6)$$

where $\lambda = e - 2 \approx 0.72$ and $\rho_Z = \max\{\sigma_Z^2, \varepsilon_r\mu_Z\}$. This stopping rule and the likelihood weighting algorithm form the foundations of both the bounded variance [Dagum and Luby, 1997] and the AA algorithms [Dagum et al., 2000]. Pradhan and Dagum [1996] tested these two algorithms on a 146 node, multiply connected medical belief network. Their results show that both algorithms are promising.

There are several differences between the algorithms proposed in this paper and the bounded variance and the AA algorithms. First, the current algorithms are based on tighter stopping rules. The Generalized Zero-One Estimator Theory does not have a relation with variance when $\sigma_Z^2 \leq \varepsilon_r\mu_Z$. So, when $\mu_Z$ is very small (this occurs often when there are many evidence nodes), the Generalized Zero-One Estimator Theory requires a prohibitive number of samples to achieve a reasonable numerical accuracy, no matter how small the variance is, since the required number of samples is inversely proportional to $\mu_Z$. Second, although the bounded variance algorithm considers the bound expressed by inequality (5), it is not tight enough and will often require a prohibitive number of samples.

Using the largest value obtained from the generated samples as bound leads to better results. Third, the AIS-BN algorithm is significantly better than the likelihood weighting algorithm — in several tested large networks with many evidence nodes we typically observed two orders of magnitude difference in accuracy [Cheng and Druzdzel, 2000]. Finally, we used different methods to construct and prove the SA-$\mu$ and SA-$\sigma$ algorithms. In the AIS-BN-$\mu$ and AIS-BN-$\sigma$ algorithms, the required number of samples is calculated dynamically based on the currently available samples.

## 6 Experimental Results

We performed empirical tests using the AIS-BN-$\sigma$ algorithm. The network used in our tests is a subset of 179 nodes of the CPCS (Computer-based Patient Case Study) network [Pradhan et al., 1994], created by Max Henrion and Malcolm Pradhan.

### 6.1 Method

We generated a total of 75 test cases with a varying number of evidence nodes (15 test cases for each: 15, 20, 25, 30, and 35 evidence nodes). The evidence was generated randomly from among those nodes that described various plausible medical findings. The least and the most likely evidence was $4.8 \times 10^{-48}$ and $7.3 \times 10^{-6}$ respectively. In over 50% of the test cases, $\Pr(\mathbf{E} = \mathbf{e})$ was less than $4.5 \times 10^{-22}$.

In each test case, we computed the posterior probabilities on each of the five disease nodes in the network using both an exact algorithm and the AIS-BN-$\sigma$ algorithm and subsequently calculated the relative error. The states of the disease nodes that we measured were "severe" and "present." There were a total of $75 \times 5 = 375$ relative error data in our test. For every posterior probability, we called function Estimate_Prob($\mathbf{W} = \mathbf{w}$) twice, obtaining a total of what we believe are 750 realistic data points for our analysis.

When we called function Estimate_Prob($\mathbf{W} = \mathbf{w}$), we generated 25,000 samples while learning the importance function (the updating interval was $l = 2,500$ samples). We did not use these samples in our estimates. Subsequently, we generated samples using the learned importance function. To avoid a possibly large estimation error, we collected at this stage at least 1,000 samples. We then continued sampling up to the estimated minimum required number of samples or 100,000, whichever was smaller.

The learning rate $\eta(k)$ used in our experiments was based on Theorem 3.1. It is a function of the ratio of the minimum required number of samples between two neighboring stages. We believe that this type of learning rate will also be suitable for other networks. If $N^{(k)}$ and $N^{(k-1)}$ are the minimum required number of samples corresponding to the sampling distributions $\Pr^{(k)}(\mathbf{X} \backslash \mathbf{E})$ and $\Pr^{(k-1)}(\mathbf{X} \backslash \mathbf{E})$ respectively, then

$$\lambda(k) := \frac{N^{(k-1)}}{N^{(k)}} = \frac{b^{(k-1)}}{b^{(k)}}.$$

This is derived from inequality (3). We used $\widetilde{b}$ instead of $\widetilde{N}$ to estimate $\lambda(k)$, as this avoided introducing another estimated value $\widetilde{\mu}_Z$. $\widetilde{b}$ was estimated using the largest value encountered in the samples. Given that typically the initialized importance sampling function was far from optimal, we let $\eta(k)$ be equal to 0.5 in the first three updating stages so that the learning algorithm had a good chance of jumping out of a possible local minimum. $\eta(k)$ used in our test can be expressed by the following formula

$$\eta(k) = \begin{cases} \frac{1}{2} & k < 3 \text{ or } \lambda(k) > 5 \\ \frac{1}{4} \cdot \log_5(5 \cdot \lambda(k)) & k \geq 3 \ \& \ \frac{1}{2} \leq \lambda(k) \leq 5 \\ 0.1423 & k \geq 3 \ \& \ \lambda(k) < \frac{1}{2} \end{cases}.$$

The above values were determined empirically based on a small number of test cases in the CPCS network that were not used in our experiments. The reason for using log function here is that we wanted to slow down the learning rate to avoid the potential oscillation when $\lambda(k)$ is large. It seems that the above learning rate performed better than the learning rate $\eta(k) = a(b/a)^{k/k_{\max}}$, used in [Cheng and Druzdzel, 2000].

Other parameters used in our test included $\varepsilon_r$=0.025 and $\delta$=0.025. Following the definition of $\delta = f_\sigma(\delta_s, \varepsilon_r)$, we obtain $\delta_s = 0.0223$, which means, according to formula (2), that the probability of estimates whose relative error is greater than 5% should be less than 5%. The remaining parameters used in our tests were identical to those reported in [Cheng and Druzdzel, 2000].

### 6.2 Results

Figure 2 shows the distribution of relative error among all tested cases with the summary data in Table 1. The percentage of estimates whose relative error was greater than 5% is 2.4%, less than 5%. We also can see that the percentage of estimates whose relative error is greater than 2.5% is not too big, only 5.9%. These results show that the estimates are still a little conservative.

Figure 3 shows the distribution of the minimum required number of samples to satisfy the precision requirement (a 2.5% relative error with 2.5% failure probability). We can see that only 2.8% data exceeded our upper limit of 100,000 on the number of samples. More than 80% of the estimates required less than 10,000 samples and almost half required less than 1,000 samples. Based on a Pentium II, 450 MHz Windows computer, the correspondence between the number of samples and the execution time in the CPCS network with 20 evidence nodes in our experiments is as follows. Learning the importance function took about 6.3 seconds. Without learning, the algorithm generated about 5,880 samples per second. So, 10,000 samples needed only about 1.7 seconds. About half of the estimates needed only 1,000 samples. As we suggested before, if after several updating stages, we find that the minimum number of samples needed to reach the required precision is still prohibitive, it may pay to restart the process with a different initialization method. In our experiment, we have tried another method. If the required number of samples was prohibitive (greater than our upper limit of 100,000 samples), we called the function Estimate_Prob($\mathbf{W} = \mathbf{w}$) again. About 60% of such cases were eliminated in the second call, i.e., a different random number seed partially solved the problem.

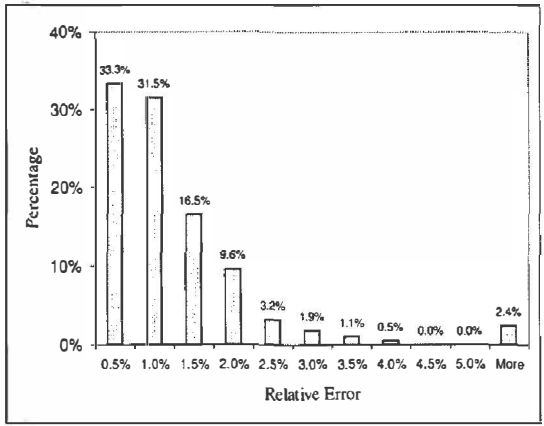

Figure 2: The distribution of relative error in the estimation of the posterior probabilities of the diseases. Total number of data points is 375.

Table 1: Summary relative error for the CPCS network. See Figure 2.

| $\mu$ | $\sigma$ | Min | Median | Max |
| --- | --- | --- | --- | --- |
| 1.1% | 1.6% | 0.00049% | 0.75% | 18.8% |

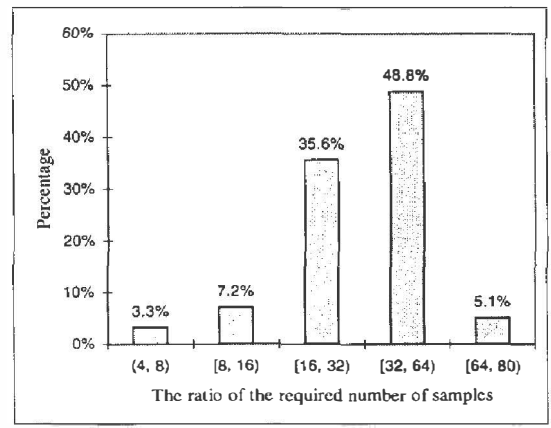

Figure 4: The distribution of the ratio of the required number of samples between the AIS-BN-$\mu$ and the AIS-BN-$\sigma$ algorithm for the CPCS network. Total number of data points is 750.

We also compared the efficiency of the AIS-BN-$\sigma$ algorithm with the efficiency of the AIS-BN-$\mu$ algorithm. Using inequality (3) and the estimated values $\tilde{b}$ and $\tilde{\mu}$, we can calculate a $\widetilde{N}$ for the AIS-BN-$\mu$ algorithm. Figure 4 shows the ratio between this number and the number obtained from the AIS-BN-$\sigma$ algorithm. We can see that AIS-BN-$\mu$ required at least four times the number of samples required by AIS-BN-$\sigma$. The maximum times were as high as 79.4 seconds. 89.5% of the test cases required over 16 times number of samples. From these data we conclude that the AIS-BN-$\sigma$ algorithm is significantly better than the AIS-BN-$\mu$ algorithm. Even if the value of $\sigma$ is estimated conservatively, it can still lead to large savings in computation.

## 7 Conclusion

We presented two algorithms — AIS-BN-$\mu$ and AIS-BN-$\sigma$ for confidence probabilistic inference in Bayesian networks. These algorithms can guarantee that the estimated results are the $(\varepsilon_r, \delta)$ relative approximation of the exact values if we know the exact values of the upper bound $b$ and the variance $\sigma^2$ of the estimated random variable. If we do not know these values, we can use the estimated $b$ and $\sigma^2$ to estimate the minimum required number of samples $N$. Although this estimation method introduces error, our experimental results show that the approximation is still reasonable and conservative. By learning the optimal importance function, sampling algorithms with the estimation algorithms can provide substantial computational savings. While they are heuristic in nature, they perform excellent in practice.

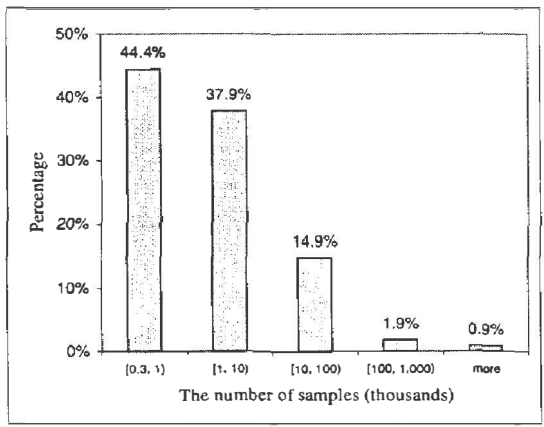

Figure 3: The distribution of the minimum required number of samples that is needed to estimate Pr($\mathbf{W} = \mathbf{w}$) with ($\varepsilon_r = 2.5\%, \delta = 2.5\%$) relative approximation. Total number of data points is 750.

Our experiments have also shown that the AIS-BN-$\sigma$ algorithm seems to be significantly better than the AIS-BN-$\mu$ algorithm and our recommendation is to adopt it in practical belief updating algorithms. Although in this paper we base the AIS-BN-$\mu$ and AIS-BN-$\sigma$ algorithms on the AIS-BN algorithm, our results are applicable to other sampling algorithms, as long as these algorithms generate independent samples.

## Acknowledgements

This research was supported by the National Science Foundation under Faculty Early Career Development (CAREER) Program, grant IRI-9624629, and by the Air Force Office of Scientific Research under grant number F49620-00-1-0112. Malcolm Pradhan and Max Henrion of the Institute for Decision Systems Research shared with us the CPCS network with a kind permission from the developers of the Internist system at the University of Pittsburgh. Jeff Schneider and anonymous reviewers provided us with useful suggestions for improving the clarify of the paper. All experimental data have been obtained using SMILE, a Bayesian inference engine developed at the Decision Systems Laboratory and available at http://www2.sis.pitt.edu/~genie.